\documentclass[letterpaper, 10 pt, conference]{ieeeconf}  % Comment this line out if you need a4paper

\IEEEoverridecommandlockouts                              % This command is only needed if 
\usepackage{amsmath} % assumes amsmath package installed
\usepackage{amssymb}  % assumes amsmath package installed
\usepackage{svg}
\usepackage{pgfplots}
\pgfplotsset{compat=1.9}
\usepackage{graphicx}
\usepackage{caption}
\usepackage{xcolor}
\usepackage{textcomp}
\usepackage{caption}
\usepackage{subcaption}
\usepackage{float}
\title{\LARGE \bf
Self-adapting Robotic Agents through Online Continual Reinforcement Learning with World Model Feedback}
\author{Fabian Domberg$^{1}$ and Georg Schildbach$^{1}$% <-this % stops a space
\thanks{$^{1}$Autonomous Systems Lab (ASL), Institute for Electrical Engineering in Medicine,
        University of Lübeck, 23562 Lübeck, Germany
        {\tt\small f.domberg@uni-luebeck.de}}%
}

\begin{document}
\bstctlcite{IEEEexample:BSTcontrol}

\maketitle
\thispagestyle{empty}
\pagestyle{empty}

%%%%%%%%%%%%%%%%%%%%%%%%%%%%%%%%%%%%%%%%%%%%%%%%%%%%%%%%%%%%%%%%%%%%%%%%%%%%%%%%
\begin{abstract}
As learning-based robotic controllers are typically trained offline and deployed with fixed parameters, their ability to cope with unforeseen changes during operation is limited. Biologically inspired, this work presents a framework for online Continual Reinforcement Learning that enables automated adaptation during deployment. Building on DreamerV3, a model-based Reinforcement Learning algorithm, the proposed method leverages world model prediction residuals to detect out-of-distribution events and automatically trigger finetuning. Adaptation progress is monitored using both task-level performance signals and internal training metrics, allowing convergence to be assessed without external supervision and domain knowledge. The approach is validated on a variety of contemporary continuous control problems, including a quadruped robot in high-fidelity simulation, and a real-world model vehicle. Relevant metrics and their interpretation are presented and discussed, as well as resulting trade-offs described. The results sketch out how autonomous robotic agents could once move beyond static training regimes toward adaptive systems capable of self-reflection and -improvement during operation, just like their biological counterparts.

Code available at: www.after-review.com/myCode

\end{abstract}

%%%%%%%%%%%%%%%%%%%%%%%%%%%%%%%%%%%%%%%%%%%%%%%%%%%%%%%%%%%%%%%%%%%%%%%%%%%%%%%%
\section{INTRODUCTION}
Learning-based control systems lack a fundamental capability when compared to their biological counterparts. That is, the ability to continually learn throughout their lifetime. Today's robots come with fixed, pre-programmed behavior that often breaks down upon a deviation from the nominal situation. While optimizing robustness through randomization and bigger datasets does have an effect, it can ultimately only go so far. Sooner or later, an unforeseen situation, referred to as an out-of-distribution event, will occur. Research suggests that (higher) biological intelligences use signals like this to drive learning. In particular, the \emph{violation-of-expectation} and \emph{minimization-of-surprise} theories state that there is an internal model responsible for gauging the familiarity with a situation at hand and, if necessary, initiating learning \cite{berseth2019smirl, margoni2024violation}. Besides enabling humans and animals to toggle between learning and executing mode, this internal model also enables planning, that is, predicting the consequences of actions without having to actually take them in the real world. In the following, we present a method that applies these theories to Reinforcement Learning (RL), exploring whether they pose useful in enabling robotic agents to autonomously adapt to changes.

% \begin{figure}[t!]
%     \centering
%     \hspace{0.3cm}\includegraphics[width=0.8\linewidth]{figs/mainfig_1.png}
%     \caption{Illustration of our proposed method. Instead of discarding a world model after training, it can be used during inference to continuously generate predictions into the future. These can then be compared to the real outcomes and thus provide a measure about the familiarity of the agent with the situation at hand. This, in turn can be used to implement universal safety checks for learning-based controllers.}
%     \label{fig:illustr}
% \end{figure}

\section{RELATED WORK}
The challenge of adapting a decision making system to changing circumstances has previously been explored in classical control theory. Here, fault detection techniques monitor system behavior for anomalies that indicate component faults. \emph{Adaptive control} methods adjust controller parameters online to cope with uncertain or time-varying system dynamics. Fault-tolerant control (FTC) integrates ideas from both to maintain controller performance despite faults \cite{blanke2001concepts}. It is often distinguished between active and passive FTC, where passive deals with anticipated faults through built-in robustness, and active monitors and reconfigures controllers online. These methods are known to be highly robust and sample efficient, and have been successfully demonstrated in many engineering applications. However, they require first principles modeling and manual design, and hence they are limited to relatively simple systems and/or environments.

In modern RL there are multiple existing propositions on how to deal with changing, a.k.a. dynamic or non-stationary, environments. However, as this is a relatively young and fast-evolving field, the community has not yet converged on a unique problem definition and terminology. Most often it is referred to as Continual Reinforcement Learning (CRL), though frequent aliases include \emph{lifelong-}, \emph{online-}, and \emph{never-ending} -learning, as well as \emph{continual-} and \emph{novelty adaption}. Additionally, there are related fields such as \emph{transfer learning}, \emph{meta-learning}, and \emph{multi-task RL}, which have overlapping aspects. In their comprehensive surveys, both Khetarpal et al. \cite{khetarpal2022towards} and Zuffer et al. \cite{zuffer2025advancements} review existing CRL approaches and attempt to formulate a taxonomy of its sub-categories. We refer the interested reader to their work and will use the term CRL in the following.

In general, there are two fundamental challenges in CRL. The first is detecting a change, the second the mechanism by which to adapt to it. The former is again known under many different terms, including \emph{context}, \emph{change-point}, \emph{domain-shift}, \emph{out-of-distribution}, \emph{novelty}, or \emph{task} -detection \cite{dick2024statistical}. Similarly, the second is sometimes also referred to as \emph{policy reconfiguration}, \emph{model updating}, or \emph{behavioral adjustment}. Beyond differences in nomenclature, there is also a diverse set of methods and algorithms in these individual areas of research. Here, we focus only on those that combine both.

An early work on CRL is by da Silva et al. \cite{da2006dealing}. They devise different partial models for each environment, each consisting of a predictor and a policy, optimized jointly for only this particular setting. When a change occurs and suddenly none of the existing predictors have sufficient accuracy, a new model duo is created. Both its predictor and policy component are then trained from scratch, exclusively on post-change datapoints. The maintenance of these specialized models allows them to circumvent one of the key ongoing challenges of the field, namely \emph{catastropic forgetting}, i.e. the tendency of such adaptive agents to lose their skill on previous tasks over time. However, it also means they have to train an entire new model duet for every new situation.
In a more recent work, Xu et al. pursue a similar approach. By utilizing a dynamically adapting mixture of Gaussian Process models, called \emph{experts}, they are able to distinguish between environments through measuring the models' prediction errors \cite{xu2020task}. Again, once a change is detected, which cannot be sufficiently modeled by the existing experts, they add a new one.
Notable are also two works by Nagabandi et al., in the first of which they also utilize a mixture-of-experts \cite{nagabandi2018deep}. In the latter, they directly apply gradient updates to a single model \cite{nagabandi2018learning}. Both essentially again attempt to update a dynamics predictor as new data come in.

Opposingly to most of these works, we maintain and continuously update a single actor and predictor, respectively. We achieve this by making use of recent advances in model based RL (MBRL), specifically the DreamerV3 algorithm \cite{hafner2025mastering}. Prior to us, Kessler et al. were able to prove its the effectiveness for this use-case, showing state-of-the-art performance in terms of knowledge transfer between tasks, reduced catastrophic forgetting, and sample efficiency \cite{kessler2023effectiveness}. Though, as with most prior works, they use simple, discrete and a-priori known environments where task boundaries are well distinguishable from one another. In contrast to their work, we focus on more complex, continuous control problems where task and environment remain largely the same, but small, potentially gradual, arbitrary changes require continuous adaption. Additionally, we also utilize the predictive \emph{world model} itself to trigger the adaptions automatically, instead of manually initiating them. Making our algorithm, to the best of our knowledge, the first fully-automated, continuous control, open-set CRL method.

% In particular, our contributions include:
% \begin{itemize}
%     \item showing how a predictive world-model can be used to detect out-of-distribution events and automatically trigger adaption to them,
%     \item exploring means of automatically gauging the completeness of an adaption process,
%     \item validating our results in contemporary RL simulation and in the real world.
% \end{itemize}

% OTHERS: often distinct task boundaries;
% OURS: take not just prediction as change-measure, but also task-performance (through reward);

\section{METHODOLOGY}
In the following, we provide details regarding the means of out-of-distribution detection and automatic fine-tuning. Our algorithms design choices are explained and relevant experiments sketched out.

\subsection{Background: DreamerV3}

Our approach builds on DreamerV3, a state-of-the-art MBRL algorithm that jointly learns a latent world model and a policy \cite{hafner2025mastering}. World model, in this case, refers to a recurrent state-space model (RSSM) of environment dynamics that allows predicting future states, values and rewards. It is trained in a supervised manner on the transitions generated by the policy as this interacts with the environment. The core idea is that the policy can largely be trained on artificial, or \emph{dreamt}, interactions, generated by the world model. This drastically reduces the amount of real environment interactions required, making training vastly more sample-efficient than model-free algorithms, such as PPO or IMPALA \cite{hafner2025mastering}.

More precisely, the RSSM world model works in the following way. Let $X_t \in \mathbb{R}^d$ denote the full observation vector at time $t$, consisting of individual state-space variables $x_{t}^{(j)}$, $j=1,\dots,d$. In the following, we use $x_t$ to refer to a single such state variable (i.e., one component of $X_t$). Given an observation $X_t$ and action $a_t$, an encoder maps the former to a stochastic latent state $z_t$, while a deterministic recurrent state $h_t$ summarizes past information. The latent dynamics are modeled as
\begin{equation*}
    h_t = f_{\theta}(h_{t-1}, z_{t-1}, a_{t-1}),
\end{equation*}
\begin{equation*}
    z_t \sim q_{\theta}(z_t \mid h_t, X_t),
\end{equation*}
where $\theta$ denotes the world model parameters. 

Future latent states are predicted using the learned transition model
\begin{equation*}
    \hat{z}_{t+1} \sim p_{\theta}(z_{t+1} \mid h_{t+1}),
\end{equation*}
which enables multi-step rollout in latent space. Predicted observations and rewards are then obtained via decoder and reward heads:
\begin{equation*}
    \hat{X}_{t} \sim p_{\theta}(X_t \mid h_t, z_t),
\end{equation*}
\begin{equation*}
    \hat{r}_{t} \sim p_{\theta}(r_t \mid h_t, z_t).
\end{equation*}

By iteratively applying this world model, imagined trajectories $(\hat{X}_{t+i}, \hat{r}_{t+i})_{i=1}^{n}$ can be generated over a prediction horizon $n$. As in the original paper, we select $n=15$, striking a balance between computational complexity, sample-efficiency and acceptable model error.

\subsection{Change Detection}
Our method is conceptually simple: A DreamerV3 model is trained as per regular on a given environment until convergence. Then, during policy inference, its world model is used to continuously predict the future $n$ states, given the current, which are then compared against the actual outcome later. The underlying assumption here is that during training of the world model, it learned to model the data distribution it was subjected to with relative accuracy. If then given any out-of-distribution data later, i.e., a previously unseen robot state during deployment, this would result in a measurable error. As world model and policy are trained on the same data, it is therefore further assumed that any situation in which the world model yields a relatively high error, is also one which the policy is not confident to act in. The theory behind this is explored in more detail in our previous work \cite{domberg2025world}. As in that work, here we compute the observation and reward error, $e_{\text{obs}_{t,x}}$ and $e_{\text{rew}_{t}}$, for any state variable $x$ at a time step $t$ as the average difference between prediction $\hat{x}_{t}$ and actual measurement $x_{t}$ over the prediction horizon $n$:

\begin{equation*}
\label{eq:pred_err}
    e_{\text{obs}_{t,x}} = \frac{1}{n} \sum_{i=1}^{n} | \hat{x}_{t+i} - x_{t+i} | ,
\end{equation*}
\begin{equation*}
\label{eq:rew_err}
    e_{\text{rew}_{t}} = \frac{1}{n} \sum_{i=1}^{n} | \hat{r}_{t+i} - r_{t+i} |.
\end{equation*}\newpage
\noindent
Averaging over all state variables' respective $e_{\text{obs}_{t,x}}$ at time $t$ yields what we will refer to as the \emph{observation prediction residual} (OPR). This, along with the reward prediction residual (RPR) $e_{\text{rew}_{t}}$, will be tracked over time to detect out-of-distribution events. Note that while there certainly are more sophisticated statistical metrics that could be used, we found this simple one to be sufficient for our needs. The interested reader is referred to Zollicoffer et al., who investigate the use of the Kullback-Leibler divergence for this purpose \cite{zollicoffer2025novelty}.

For this study, we define an out-of-distribution, or change, event as a deviation from the nominal value in at least one of OPR and RPR. Since both metrics remain largely flat over time for a fully trained world model, we found it sufficient to apply a threshold-based check to detect any abnormal situation. In our experiments, we therefore compute a rolling mean and standard deviation for both metrics, classifying anything beyond three standard deviations from the mean as a change. The reason for monitoring both reward and observation residuals is that the former alone may only provide sparse and/or delayed signal, while the latter alone does not capture a change's effect on task performance. Similarly, the combination of these two is what is required for automated evaluation of the adaption progress, as explained below.

\subsection{Automatic Adaption}
Once an out-of-distribution event is detected, we start finetuning world model and policy using DreamerV3's regular training loop. That is, the robot continues to operate, collecting state-transitions and rewards to refit the world model, which is in turn used to adapt the policy in its imaginary latent space. We use the original authors implementation with the \emph{medium} model size preset ($12$ million parameters) and a \emph{train ratio} of $16$. Other than this, we do not alter any (hyper-)parameters. Note that transitions prior to a change event are not taken into the replay buffer for finetuning, as these potentially exhibit different dynamics.

To automatically determine whether and when adaption has been successful, we examine multiple different measures and criteria. Among all of these, one necessary condition sought is convergence. Just like a human expert would, our proposed system thereby checks not only whether a measure reaches a desired value once, but whether it does so reliably over time, without excessive fluctuations or unexpected trends, which would signal an ongoing learning process. Measures taken into account include OPR and RPR, introduced before, as well as algorithm internal learning signals, such as losses. As DreamerV3 optimizes both world model and policy simultaneously, there are multiple measures that have to be considered to determine its learning advance. Most importantly, it has to be taken into account that these two components are strongly interconnected. For example, since the policy is trained entirely within the world model, it may signal saturation early, as the world model is not yet able to represent the more complex situations of the environment. Similarly, the other way around, the world model might start to indicate convergence, only for the policy to eventually start exploring a yet under-represented part of the state-space. To account for this strong interconnectedness, we particularly monitor the following key metrics throughout our experiments.

\begin{enumerate}

\item \textbf{Dynamics Loss:} measures how accurately the world model can predict how the state-space evolves over time. It reflects whether the internal representation of environment dynamics is stable and consistent. A decreasing and eventually stable value indicates that the model has formed a reliable representation, while continued trends suggest that it is still adapting.

\item \textbf{Advantage Magnitude:} quantifies the strength of the policy improvement signal, i.e., how much better the actor's selected actions are compared to the critic’s prediction. High values indicate that substantial behavioral improvements are still being identified. Stabilization at a low but non-zero level suggests incremental refinement, whereas persistent decreases indicate ongoing policy adaptation.

\item \textbf{Value Loss:} indicates how accurately the model estimates the long-term outcomes of its decisions. It reflects the consistency between predicted and realized returns under the current policy. Convergence is characterized by stable fluctuations around a constant level.

\end{enumerate}

\section{RESULTS}
\label{results}
Our results comprise experiments using three different robotic systems, showcasing the general applicability of the proposed method. First, we establish a proof-of-concept on an established continuous control RL problem. Second, we apply it to common, industry-grade robot in simulation. And finally, third, we validate its effectiveness in the real-world.

\subsection{DMC Walker}
\begin{figure}[h]
    \centering
    
    \begin{subfigure}{\columnwidth}
        %\centering
        \hspace{0.03em} % Negative value to shift left, positive to shift right
        \includegraphics[width=0.983\columnwidth]{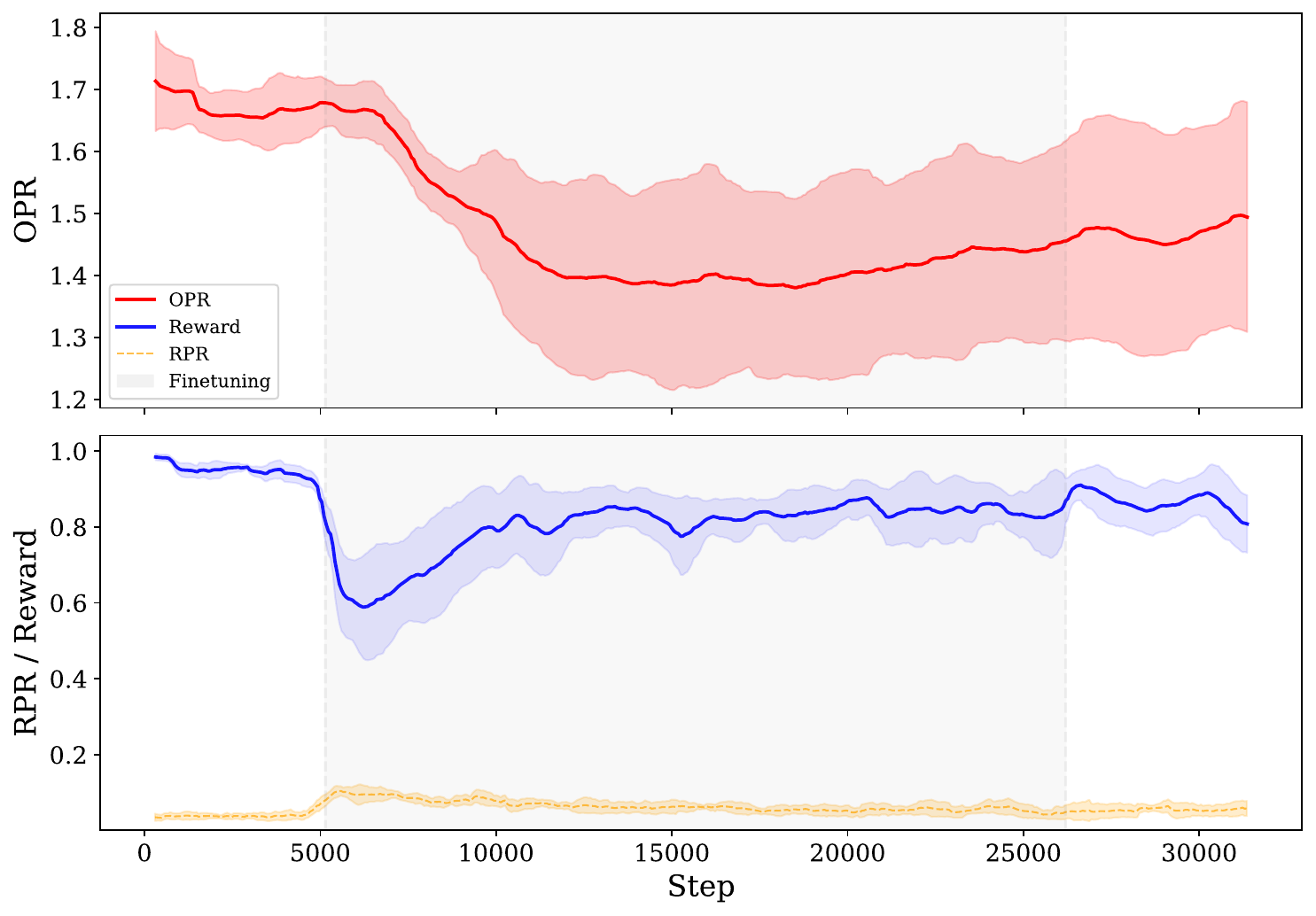}
        %\includesvg[width=0.983\columnwidth]{figs/finetune_walker}
        \caption{}
        \label{fig:dmc_walker_opr}
    \end{subfigure}
    
    \vspace{0.5em}
    
    \begin{subfigure}{\columnwidth}
        \centering
        %\includesvg[width=\columnwidth]{figs/adv_dyn_value_loss_walker}
        \includegraphics[width=\columnwidth]{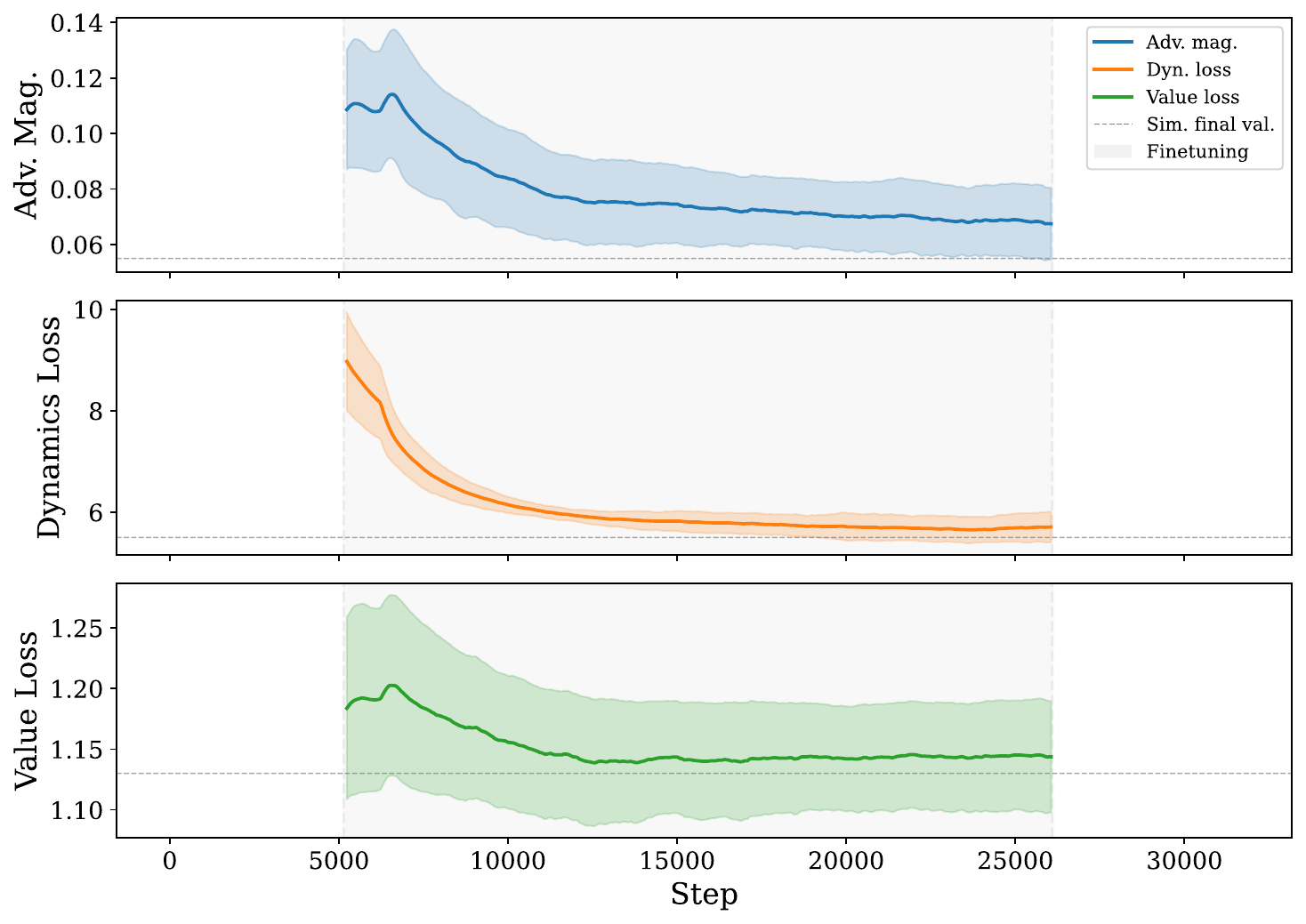}
        \caption{}
        \label{fig:dmc_walker_adv}
    \end{subfigure}
    
    \caption{Experiment on DMC's Walker Walk problem. At $5,000$ steps a random joint’s gear ratio is reduced, causing the Walker to lose its balance repeatedly. The proposed method detects the change and initiates finetuning, leading to recovery. Averaged over ten runs.}
    \label{fig:dmc_walker_comb}
\end{figure}

The \emph{DeepMind Control Suite} is an established collection of continous control RL problems \cite{tassa2018deepmindcontrolsuite}. From it, we chose the well-known \emph{Walker-walk} one, where a two-dimensional stick-figure-humanoid is tasked with keeping its body upright and walking forward. After an initial period of running the unchanged simulation, we randomly select one of the figurine's joints and half its gear ratio, simulating actuator damage. Figure \ref{fig:dmc_walker_opr} shows the responses of OPR, RPR, and reward, while Figure \ref{fig:dmc_walker_adv} depicts advantage magnitude, dynamics loss, and value loss. Mean and standard deviation are calculated over ten runs. Immediately after a joint is damaged, at $5000$ steps, reward quickly drops noticeably and RPR rises accordingly, as the Walker cannot maintain its balance. Our method recognizes this rapidly, initiating the adaption process. Within less than $10,000$ steps, i.e., $2$ minutes of simulation time, most metrics are asymptotically moving toward their pre-modification values. The Walker is back on its feet, reaching, on average, slightly less reward than before, as it still occasionally trips. The only metric with continued variability is the OPR, which is due to varying effects of random modification. That is, sometimes the Walker can continue walking nearly unaffected, whereas other times it is heavily impaired.

\subsection{Quadruped Robot ANYmal}
\begin{figure}[b!]
    \centering
    
    \begin{subfigure}{\columnwidth}
        \hspace{-0.4em}
        %\includesvg[width=0.963\columnwidth]{figs/finetune_dog}
        \includegraphics[width=0.963\columnwidth]{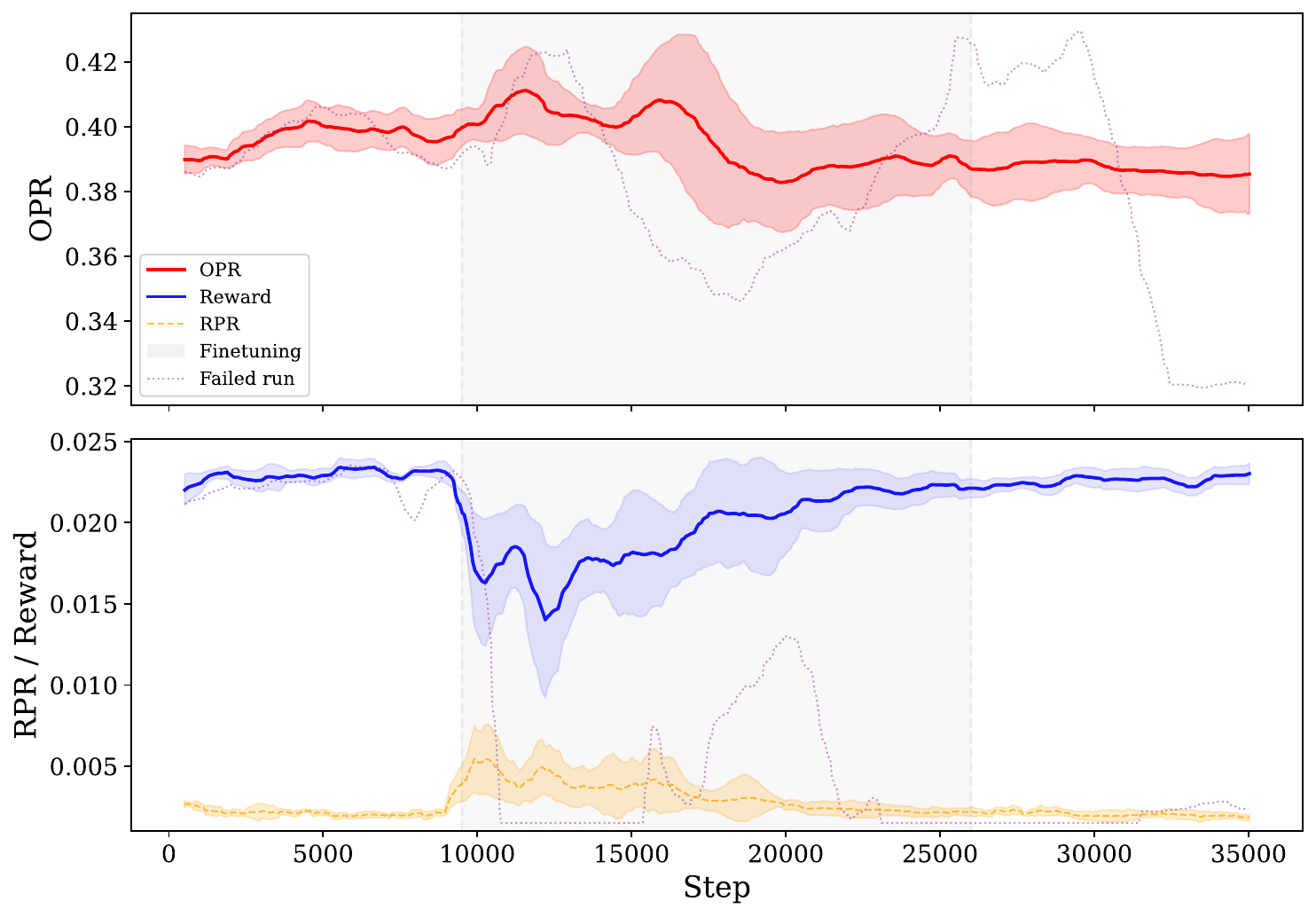}
        \caption{}
        \label{fig:anymal_opr}
    \end{subfigure}
    
    \vspace{0.5em}
    
    \begin{subfigure}{\columnwidth}
        \centering
        %\includesvg[width=\columnwidth]{figs/adv_dyn_value_loss_plot}
        \includegraphics[width=\columnwidth]{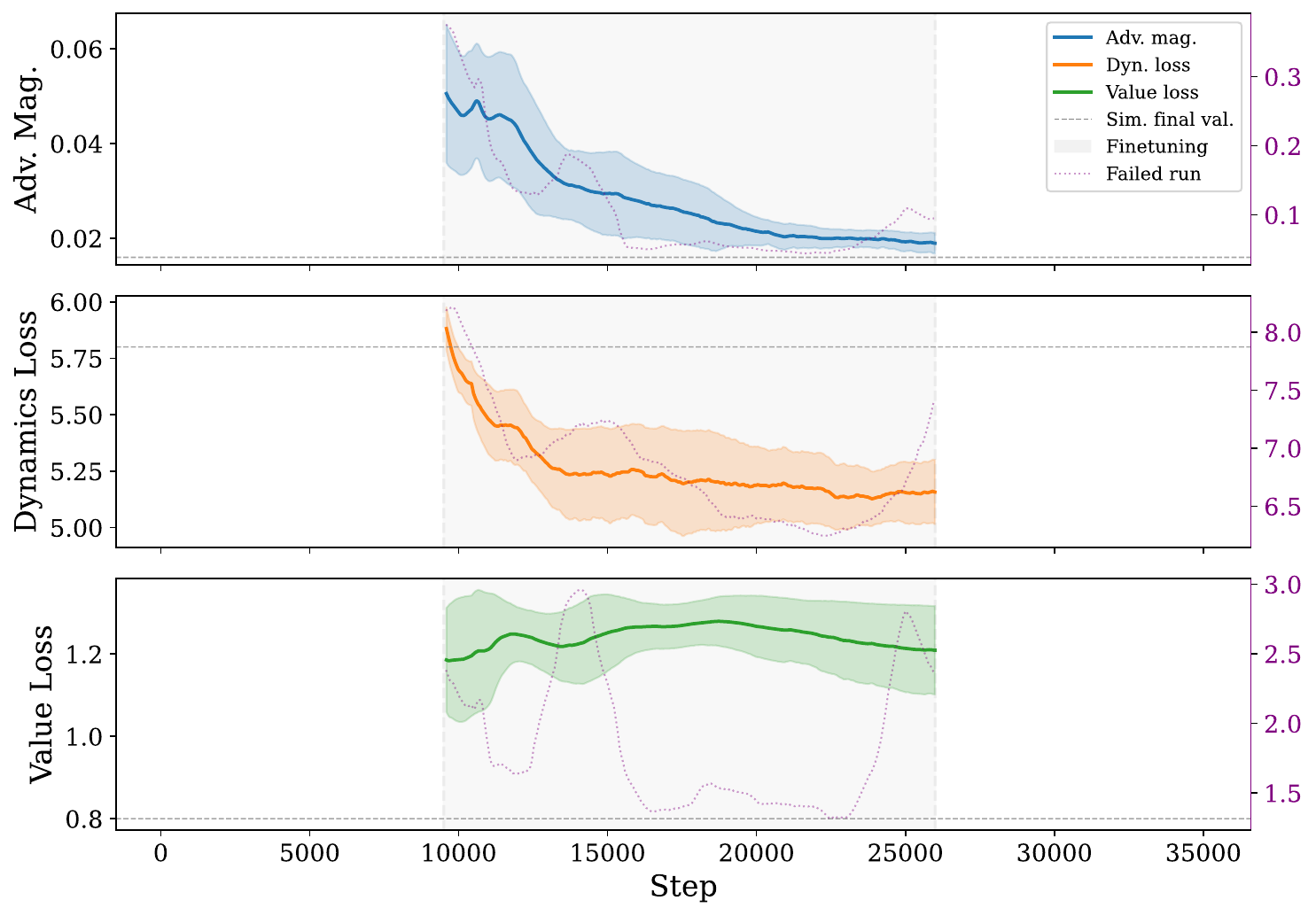}
        \caption{}
        \label{fig:anymal_adv}
    \end{subfigure}
    
    \caption{Simulated experiment using quadruped robot \emph{ANYmal}, averaged over nine runs. At $9,000$ steps, velocity limits of the right hind leg actuators are reduced, resulting in unstable locomotion and decreased reward. Our method detects the change and initiates finetuning, restoring stable walking. A failed run additionally illustrates non-convergent behavior.}
    \label{fig:anymal_comb}
\end{figure}

In this experiment, we seek to show our method's applicability to a more complex, practical use-case. We therefore choose the ubiquitous robot dog example. In particular, we select the NVIDIA Isaac Lab simulation environment, which already comes with multiple pre-configured robotic environments \cite{mittal2025isaaclab}. Among them, ones for the \emph{ANYmal} quadruped platform, used throughout both research and industry. The robot is trained, until convergence, to walk at a commanded pace in a commanded direction for $25$ million steps. Afterwards, we modify its right hind leg's three actuators to retain only one third of their previous velocity, simulating an actuator failure. Figure \ref{fig:anymal_opr} shows the resulting behavior of reward, OPR, and RPR over time, with the modification occurring at timestep $9,000$, averaged over nine runs. The robot cannot maintain its smooth motion, repeatedly tripping and occasionally falling over. Our method quickly recognizes the change as a significant drop in reward and initiates finetuning. After, on average, $5,000$ steps or four minutes, the robots' walking cycle stabilizes. Eventually, the last run stops its finetuning at $26,000$ steps. Figure \ref{fig:anymal_adv} shows the corresponding DreamerV3 internal metrics throughout the finetuning period. Again, for the initial ca. $5,000$ steps, they show elevated levels. In particular, high advantage magnitude indicates increased behavioral exploration, declining once behavior settles. Dynamics loss decreases even below its prior level. Value loss is elevated throughout the finetuning phase, only beginning a downtrend once both policy and world model have settled.

Both figures additionally show a failed run, where metrics never stabilize and thus adaption is aborted eventually. This clearly shows the distinct convergence pattern that enables automatic evaluation.

\subsection{Real World}
\label{real_car}
\begin{figure}[t]
    \centering
    
    \begin{subfigure}{\columnwidth}
        %\hspace{0.01em} % Negative value to shift left, positive to
        %\includesvg[width=\columnwidth]{figs/finetune_sim_real_socks}
        \includegraphics[width=\columnwidth]{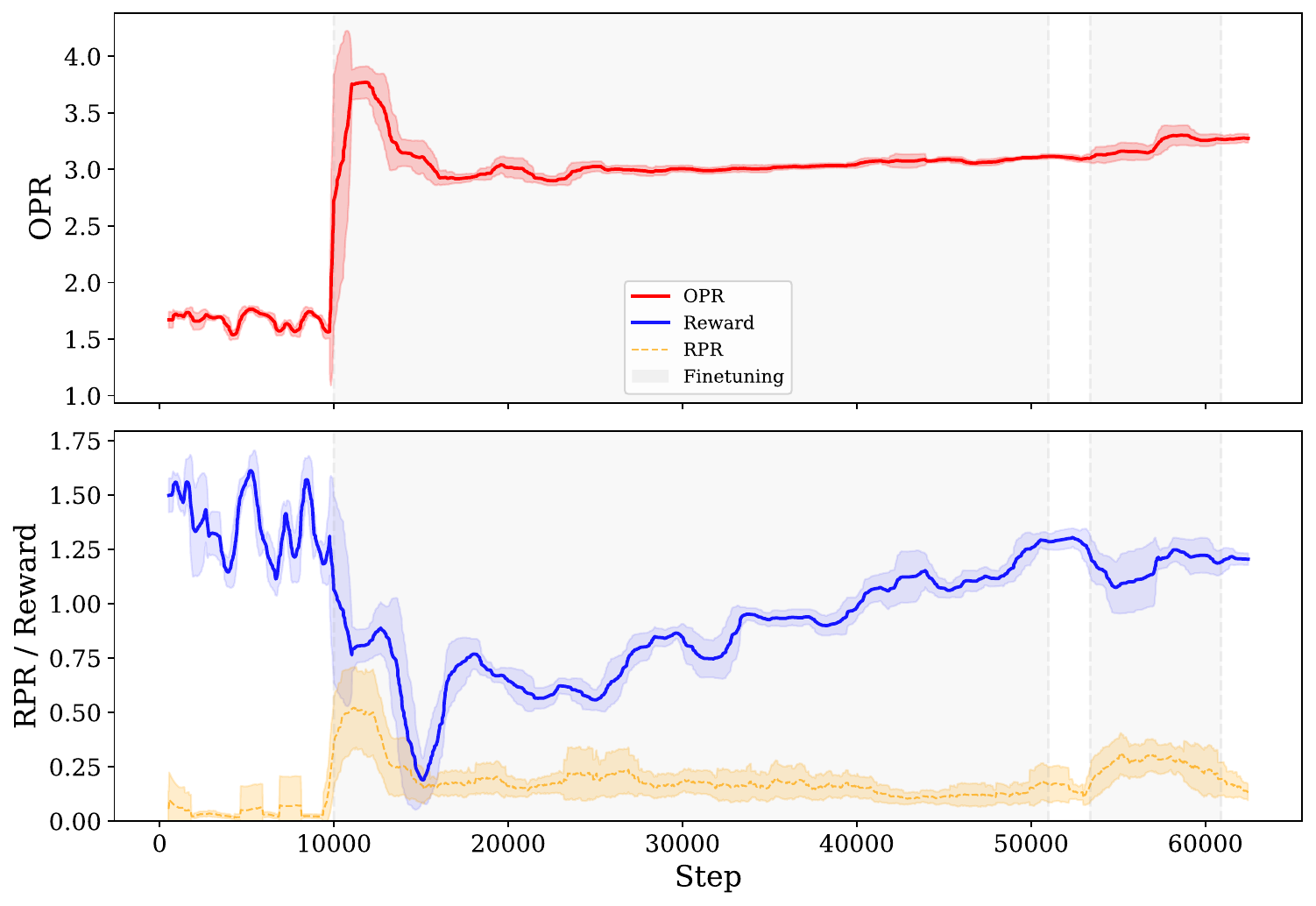}
        \caption{}
        \label{fig:real_opr}
    \end{subfigure}
    
    \vspace{0.5em}
    
    \begin{subfigure}{\columnwidth}
        \centering
        %\includesvg[width=\columnwidth]{figs/adv_dyn_loss_plot}
        \includegraphics[width=\columnwidth]{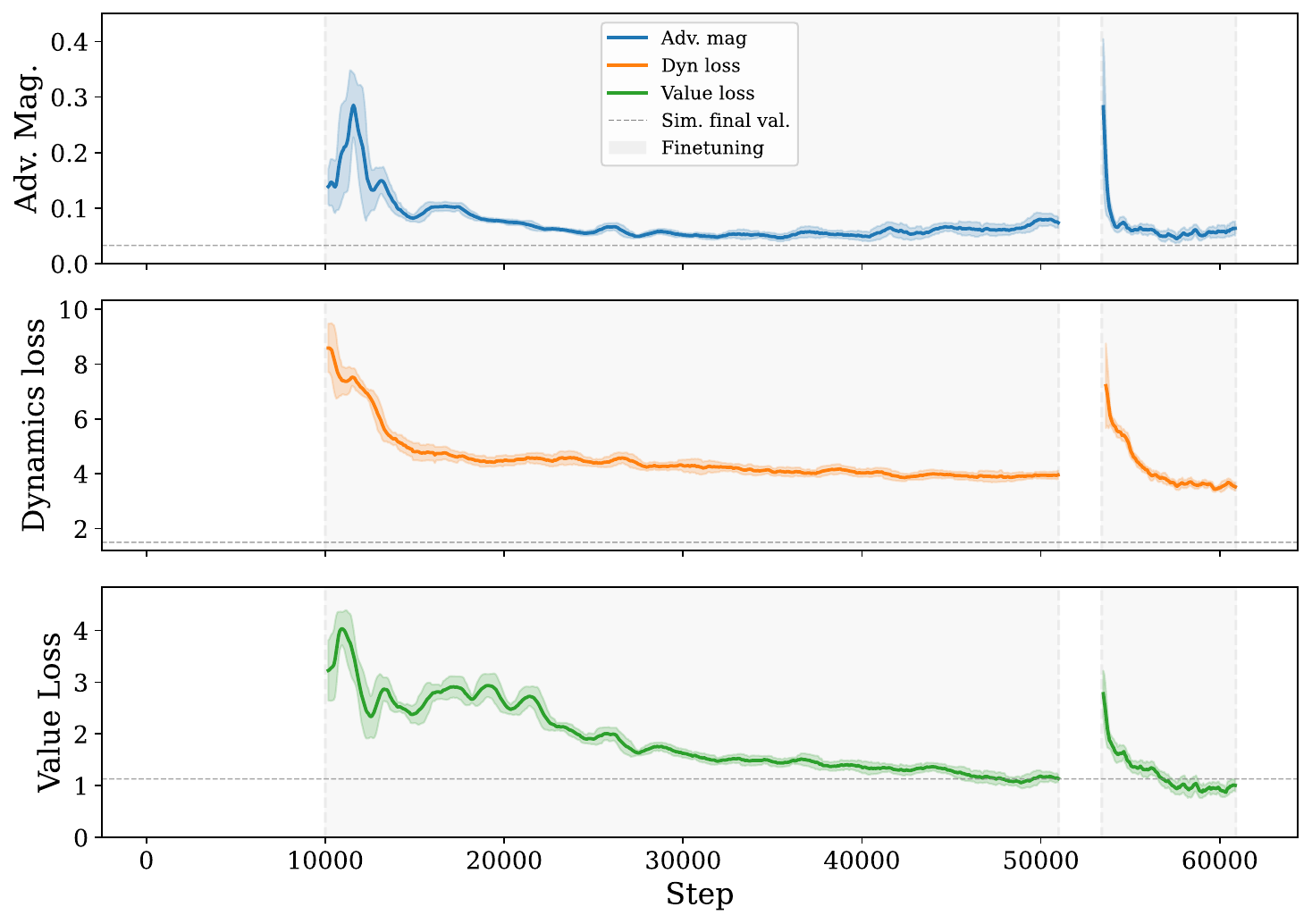}
        \caption{}
        \label{fig:real_adv}
    \end{subfigure}
    
    \caption{Experiment moving a trained model from simulation to a real, $1$:$10$ scale vehicle. At $10,000$ steps the policy is transferred to the car, leading to a surge in prediction residuals and decreased reward. Finetuning stabilizes behavior and slowly recovers. After $50,000$ steps, rear-wheel friction is reduced, causing a secondary adaption phase and subsequent recovery.}
    \label{fig:real_comb}
\end{figure}

Finally, to validate our results, we conduct a real-world experiment using a $1$:$10$ scale model car based on the F1Tenth project \cite{OKelly2019F1TENTHAO}. Before moving to the real car, we initially train world model and policy in the associated simulation. In addition to being standard procedure, this serves as a source for a first out-of-distribution event, as the simulation does not match the real-world perfectly. The reward function is kept simple, linearly rewarding forward velocity and penalizing collisions with the track boundary. After the initial training for $10$ million simulation steps, we move the model onto the real car in our lab, running at $20$ Hz, and observe its response. Figure \ref{fig:real_opr} shows the resulting curves for OPR, RPR, and reward, where timestep $10,000$ marks the sim-to-real transition. Immediately, OPR surges, while reward drops soon after. The vehicle exhibits more jerky driving than in simulation and frequently gets too close to the walls and ultimately crashes. Our method again quickly recognizes this abnormality and initiates finetuning. During the next circa $10,000$ steps, i.e., $8$ minutes of wallclock time, the car's behavior stabilizes. Looking at the corresponding internal model metrics declining, depicted in Figure \ref{fig:real_adv}, confirms this. Afterwards, learning continues on a more steady path. From $20,000$ steps onward, the vehicle essentially only optimizes already existing behavior, such as increasing its speed. This is apparent from the steady, slightly down-trending advantage magnitude curve. As this slightly starts to grow again, around $40,000$ steps, the vehicle can no longer improve much by simply increasing speed and therefore can be seen trying to attain more reward through different means, such as taking corners more tightly and smoothing its steering. Shortly after $50,000$ steps, the reward reaches simulation-like levels and our algorithm ends the fine-tuning phase, given that all other metrics have largely converged back also. The remaining difference in OPR and dynamics loss, which essentially measure the same thing, is likely explained through unmodeled differences between simulation and reality.

After adaption from simulation to the real car is done, another experiment is conducted immediately after. At around timestep $52,000$ we put socks on the vehicle's rear wheels, reducing their friction with the ground. Following this, the reward drops by around $20$\% as the car occasionally slips and spins out of control when cornering. The effect on OPR remains low, only being noticeable through an increase in standard deviation. Though looking more closely at individual state variables' logs, we find that there is indeed a significant peak in the vehicle's angular velocity prediction. Yet, the overall effect on OPR low, likely due to averaging. Apart from this, there are no noticable events throughout this second finetuning phase. Losses quickly drop down to pre-socks levels as the policy settles on driving slightly slower, as not to start slipping, resulting in an accordingly slightly lower total reward.

\section{DISCUSSION \& FUTURE WORK}
Although our experiments show promising results, they are also indicative of a few caveats and open questions regarding the proposed method.

\subsection{Automatic Adaption}
As apparent from the experiments conducted throughout Section \ref{results}, it is indeed feasible to automatically judge an adaption process. However, it is important to note that there is no singular measure for this, but requires the consideration of multiple key metrics and is not guaranteed to succeed. The point at which adaption is considered complete at all is subject to debate. This is because of the different use-cases and their resulting trade-offs. For example, a quadrupedal inspection robot in an industrial facility that suffers a joint fault may only need to adapt sufficiently to reach a safe zone and wait for a mechanic. Here, the environment is sensitive, and further experimentation could worsen the damage or endanger surrounding equipment. In this case, adaptation is conservative and may justifiably stop once safety is ensured. In contrast, an autonomous rover on another planet cannot rely on external help. If it gets stuck, it must continue adapting, even at the risk of additional strain or other high cost. In general, trade-offs can include the degree of autonomy, the surrounding environment, as well as time and/or computational resources. Among all criteria, reward is going to be among the most meaningful, because, ideally, it boils down a systems goals and tradeoffs into a single quantity \cite{Silver2021RewardIE}. Though, as in reality this is often difficult to achieve, we recommend the use of task-specific criteria and acceptance thresholds, such as whether or not a robot can maintain its balance, or stay within certain other minimum safety bounds. If simple gauges like this are not sufficient, one could also consider the use of a language-model-like system to assess the situation \cite{Ma2023EurekaHR}.

\subsection{Prior Skill Retention}
This is one of the central questions in Continual Reinforcement Learning. Known as the \emph{stability-plasticity dilemma}, there is usually a trade-off between how quickly new skills are attained versus old ones being lost \cite{Grossberg1987TheAO}. Kessler et al., explore different strategies for sampling from a lifelong experience replay buffer, showing different trade-off equilibria \cite{kessler2023effectiveness}. Other than them, and most other CRL research, we deliberately focus on an open, incrementally evolving set of possible situations that need adaption to, not actively trying to retain any, potentially outdated, prior knowledge. In other words, we assume that anything in a given robotic system's operational setting can change, including towards something that contradicts prior truths. An example of this would be our experiment from Section \ref{real_car}, where we modify the friction of the car's rear tires. An algorithm that retains and continues to train on this outdated experience might continue driving around corners too fast for longer. While with this stance we potentially lose adaption efficiency for situations that were experienced previously, we do gain general applicability and open-endedness. Changes whose scope are known a-priori, expected to be frequently or even cyclically encountered, can be handled more efficiently through multi-task learning.

\subsection{Great Changes Require Great Adaption}
As observable throughout our experiments, the amount of steps required for adaption is proportional to the \emph{distance} between the prior and posterior situation. For example, adapting from simulated to real car in Section \ref{real_car} takes around $40,000$ steps, whereas adapting to an isolated change in the real car's dynamics requires merely $10,000$ steps. This leads us to believe that there is no fundamental reason why this method would not be able to adapt to any arbitrary change, given enough time and an appropriate learning algorithm. To improve upon adaption speed, future work may look at the effect of already including changes during the training phase, even feeding in measures such as world model error directly, potentially somehow preparing the model for downstream non-stationary. Further, one may investigate whether adaption performance somehow relates to model size and training ratio. DreamerV3's original paper already shows sample-efficiency scaling favorably for both. As there is a trend towards larger, generalist multi-task \emph{foundation models} for RL \cite{hansen2023td, Team2024OctoAO, Brohan2023RT2VM}, our method could eventually enable their sample-efficient on-the-job finetuning. Another interesting research avenue here would be to consider changes to the reward function also. This would enable to efficiently reskill agents in the same environment, for example to purposefully adopt a drifting behavior on the scale car \cite{domberg2022deep}.

\subsection{Safety Concerns}
While our results and their implications are promising, there remain critical safety concerns, especially with real-world deployment. Afterall, RL techniques still require making a mistake multiple times before they can learn not to repeat it. Such unconstrained exploration and resulting unpredictable behavior is most often a dealbreaker. In cases where basic, rule-based safety measures are not enough, future work may, for example, explore the integration with Safe RL methods, potentially drastically reducing the number of samples of a mistake needed to extract a learning signal \cite{hafner2025mastering}. Going even further, integration with supervisory Model Predictive Control methods could theoretically entirely eliminate the need to execute an action leading to mistake at all \cite{Nezami2021ASC}.

\section{CONCLUSION}
This paper introduced a framework for enabling robotic agents to autonomously detect and adapt to changes during deployment using online learning by merging theories of biological learning from neuroscience with state-of-the-art RL. By monitoring world model prediction residuals, the system identifies these changes and initiates finetuning without manual intervention. The combination of task performance metrics and internal training signals provides a practical mechanism to judge convergence and automatically terminate adaptation. Experiments in simulation and on a physical platform show that the approach generalizes across different robotic systems and types of disturbances, including actuator degradation, sim-to-real transfer, and physical distortions. In each case, the agent restores stable behavior quickly through continued interaction with the environment. While the results indicate that MBRL provides a promising foundation for self-adapting robotic systems, several open challenges remain, including safety during adaptation, efficiency under large distribution shifts, and long-term skill retention. Further, there arise certain trade-offs for different use-cases. Addressing these aspects will be essential for deploying such systems in safety-critical or highly dynamic environments, such as the real world. We nevertheless believe that this method represents a foundational step towards autonomous, self-improving robotic agents.

\addtolength{\textheight}{-10.5cm}   % This command serves to balance the column lengths

%%%%%%%%%%%%%%%%%%%%%%%%%%%%%%%%%%%%%%%%%%%%%%%%%%%%%%%%%%%%%%%%%%%%%%%%%%%%%%%%

%%%%%%%%%%%%%%%%%%%%%%%%%%%%%%%%%%%%%%%%%%%%%%%%%%%%%%%%%%%%%%%%%%%%%%%%%%%%%%%%



%%%%%%%%%%%%%%%%%%%%%%%%%%%%%%%%%%%%%%%%%%%%%%%%%%%%%%%%%%%%%%%%%%%%%%%%%%%%%%%%

\bibliography{bibl} %Prints bibliography
\end{document}